\documentclass{article}
\usepackage[english]{babel}
\usepackage[square,numbers]{natbib}
\usepackage[preprint]{style_file}
\usepackage[utf8]{inputenc} 
\usepackage[T1]{fontenc}    
\usepackage{hyperref}       
\usepackage{url}            
\usepackage{booktabs}       
\usepackage{amsfonts}       
\usepackage{nicefrac}       
\usepackage{microtype}      
\usepackage{xcolor}         
\usepackage{graphicx}
\usepackage{caption}
\usepackage{subcaption}
\usepackage{multirow}
\usepackage{enumitem}
\usepackage{algorithm}
\usepackage{algpseudocode}
\usepackage{wrapfig}
\usepackage{amsmath}

\title{Recipes for Pre-training LLMs with MXFP8}

\author {Asit Mishra, Dusan Stosic, Simon Layton, Paulius Micikevicius \\
NVIDIA \\
Corresponding authors: \texttt{asitm@nvidia.com, dstosic@nvidia.com}
}

\begin{document}
\maketitle

\begin{abstract}

Using fewer bits to represent model parameters and related tensors during pre-training has become a required technique for improving GPU efficiency without sacrificing accuracy. 
Microscaling (\texttt{MX}) formats~\cite{ocp} introduced in NVIDIA Blackwell generation of GPUs~\cite{blackwell}
represent a major advancement of this technique, making it practical to combine narrow floating-point data types with finer granularity per-block scaling factors. In turn, this enables both quantization of more tensors than previous approaches and more efficient execution of operations on those tensors.

Effective use of \texttt{MX}-formats requires careful choices of various parameters. In this paper we review these choices and show how \texttt{MXFP8-E4M3} datatype and a specific number conversion algorithm result in training sessions that match those carried out in \texttt{BF16}. We present results using models with up to 8B parameters, trained on high-quality datasets of up to 15T tokens.

\end{abstract}

\section{Introduction}

Scaling the number of parameters of deep learning models, especially large generative models, has intensified the need for more efficient compute and memory solutions. Reducing the number of bits to represent a data type (sometimes referred to as reducing precision) is a successful strategy to improve performance while lowering memory requirements. However, maintaining model accuracy while reducing precision remains a key challenge.

Microscaling (\texttt{MX}) formats, specified by an Open Compute Project working group (OCP)~\cite{ocp}, combine narrow floating-point types with fine-grained scaling factors to deliver a balance of dynamic range, precision, and hardware efficiency. OCP v1.0 specification~\cite{ocp} and research works like~\cite{rouhani2023microscalingdataformatsdeep,rouhani2023sharedmicroexponentslittleshifting,dai2021vsquantpervectorscaledquantization} have shown a fine-grained scaling approach  
to be an effective strategy compared to granular (e.g., per-tensor) scaling approaches, especially in sub-8-bit pre-training regimes. 

NVIDIA's latest GPU generation, Blackwell~\cite{blackwell}, adds native \texttt{MX} data type support to Tensor Cores. Blackwell supports \texttt{MX} data for 8-bit (\texttt{MXFP8}), 6-bit (\texttt{MXFP6}) and 4-bit (\texttt{MXFP4}) floating-point types. 
This paper focuses on \texttt{MXFP8} pre-training of large language models (LLMs), demonstrating two key factors needed for successful pre-training (no accuracy degradation compared to \texttt{BF16} sessions):
\begin{itemize}
    \item First, \texttt{MXFP8-E4M3} datatype should be used for all tensor types, including activation gradients.
    \item Second, careful consideration of rounding is needed when converting from high-precision formats to \texttt{MXFP8}, resulting in an algorithm different from the one suggested in OCP v1.0 specification.
\end{itemize}

For each design choice, we present ablation studies on small models (e.g. 100s of millions of parameters) pretrained on short (e.g. 100s of billions) token horizons. We then apply the findings to large multi-billion-parameter models trained on trillion token horizons.

\section{Microscaling format support in NVIDIA Blackwell}
\label{sec:mxformat}

\begin{figure*}[htb]
    \centering
    \includegraphics[width=0.7\textwidth]{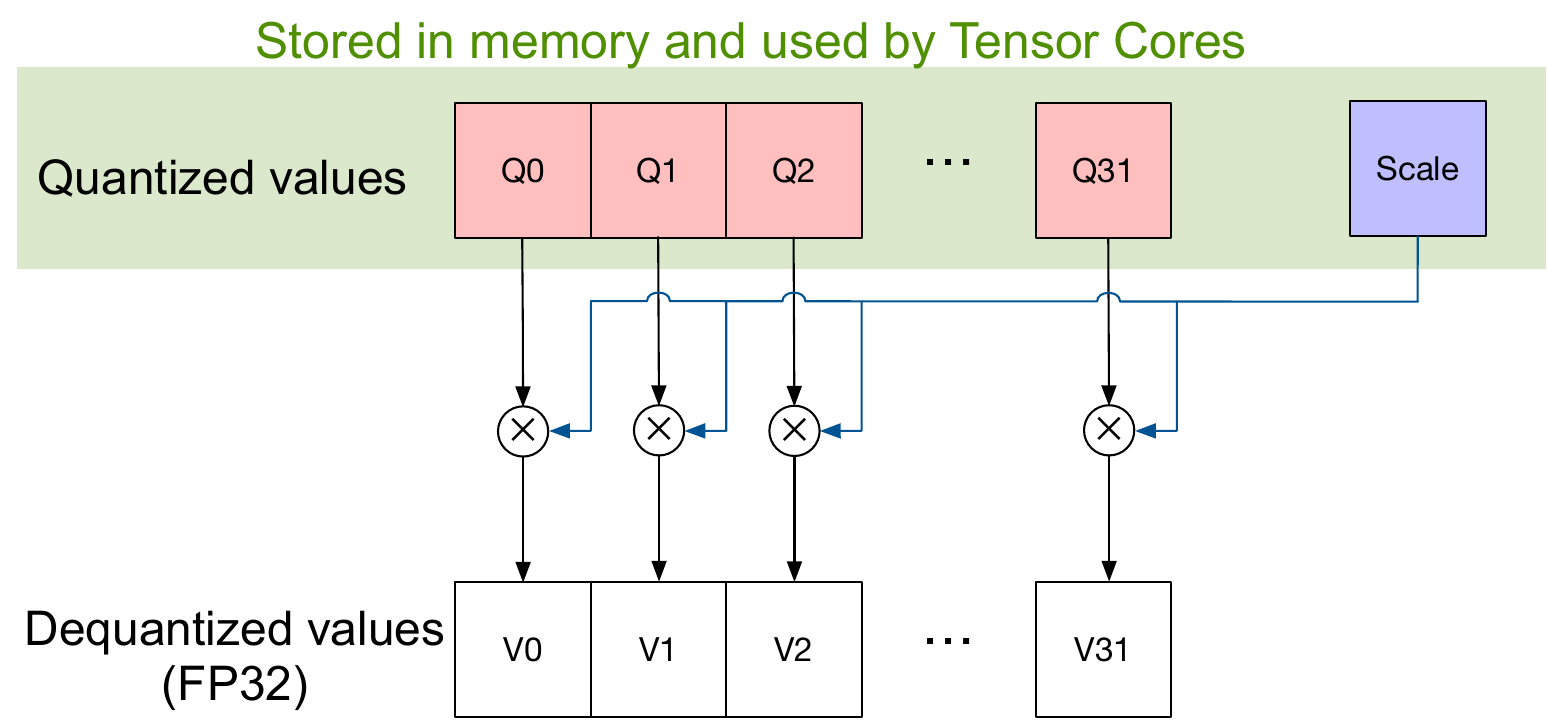}
    \caption{A single \texttt{MXFP} block (in green box) and interpretation of \texttt{MXFP} format.}
     \label{fig:fig_mxfp}
\end{figure*}

\noindent \textbf{Background:} An \texttt{MX}-format is specified by the block size \verb+K+, a shared scaling factor per block, \verb+X+, and the data-type of elements in the block. A block is a contiguous sequence of \verb+K+ elements. \verb+K+ = 32 for all \texttt{MX} types. The data-type of \verb+X+ is \verb+UE8M0+ 
and \verb+X+ \textit{encodes} either \texttt{NaN} or any power-of-two\footnote{Power-of-two is a number of the form $2^n$ where n is a positive integer, negative integer, or zero.} value in the range $2^{-127}$ to $2^{127}$. 

Given \verb+K+ input elements in a source-format, $V_i$ (typically \texttt{FP32}); $1 \leq i \leq K$, conversion to \texttt{MX}-format consists of computing \verb+X+ and $Q_i$ such that $Q_i \times$\verb+X+ is the decoded value corresponding to $V_i$. 
\verb+X+ and $Q_i$ are stored in memory instead of $V_i$. Thus, \verb+X+ serves as the decoding scale factor to decode quantized values back to their high-precision counterparts. This is depicted in Figure~\ref{fig:fig_mxfp}. 

A tensor in source-format is sub-divided into blocks of \verb+K+ elements
and converted into \texttt{MX}-format to be stored in memory and/or processed by math units in hardware.
Tensor Cores in Blackwell consume \verb+X+ and $Q_i$ to compute the dot product of two \texttt{MX}-formatted blocks. 
If the accumulated output of a dot-product operation in Tensor Cores is \texttt{FP32}, then this output is thereafter quantized to \texttt{MX}-format if a subsequent operation consuming the output needs it in that format. 
The conversion process is, $Q_i = \verb+Quantize_to_fp8+(V_i / X)$.
Section~\ref{sec:training} describes the conversion details. Fine-grained scaling helps each \texttt{MX} block independently align to the needed range of input values before quantization.

\begin{table}[htbp]
\setlength{\tabcolsep}{8pt}
\centering
\caption{\texttt{MX}-format support in Blackwell}
\label{tab:mxformat}
\begin{tabular}{@{}lllllll@{}}
\toprule
{\textbf{Format}} & {\textbf{Data}} & {\textbf{Max.}} & {\textbf{Min.}} & {\textbf{Binades}} & {\textbf{Relative speed}} \\ 
&  {\textbf{type}} &  {\textbf{normal}}  &  {\textbf{subnorm.}}  &   & \textbf{vs BF16 (Tensor} \\
& & & & & \textbf{Core math)} \\
\midrule
\multirow{2}{*}{MXFP8}  & E4M3  & 1.75 * $2^8$  & $2^{-9}$  & 17.8 & 2x \\
                        & E5M2   & 1.75 * $2^{15}$ & $2^{-16}$ & 31.8 & 2x \\
                        \cmidrule(l){2-6}
\multirow{2}{*}{MXFP6}  & E2M3  & 1.875 * $2^2$ & $2^{-3}$   & 5.9 & 2x \\
                        & E3M2  & 1.75 * $2^4$  & $2^{-3}$   & 8.8 & 2x \\
                        \cmidrule(l){2-6}
MXFP4                   & E2M1  & 1.5 * $2^2$   & $2^{-1}$   & 3.6 & 4x \\
\bottomrule
\end{tabular}
\end{table}

Table~\ref{tab:mxformat} shows the \texttt{MX}-formats supported in Blackwell. The data type column uses the convention \verb+ExMy+ to denote \verb+x+-bit for floating-point exponent and \verb+y+-bit for mantissa. For a fixed bit-width, floating-point numbers trade-off exponent width with mantissa width. Thus, \texttt{MXFP8} \texttt{E4M3} is a 8-bit data type with 1 sign bit, 4-bit for FP exponent and 3-bit of mantissa. Similarly, \texttt{E5M2}  is an 8-bit type with 5-bit for exponent and 2-bit for mantissa. Compared to \texttt{E4M3}, \texttt{E5M2}  can represent a larger dynamic range, but with less precision.
\texttt{E5M2} follows IEEE 754 conventions~\cite{ieeefp} for representation of special values, whereas the remaining data types  extend the dynamic range by not representing \verb+Infinity+ and \texttt{NaN} (\texttt{E4M3} has only one bit-pattern for \texttt{NaN}~\cite{micikevicius2022fp8formatsdeeplearning}). More bits in the exponent field translates to a larger range while more bits in the mantissa field translates to more precision within a given range.
Every floating-point type has a dynamic range that it can represent — we denote this range in terms of \textit{binades} which is the log$_2$-ratio of the maximum to the minimum finite representable in that format.

\begin{figure*}[hbpt]
    \centering
    \includegraphics[width=1\textwidth]{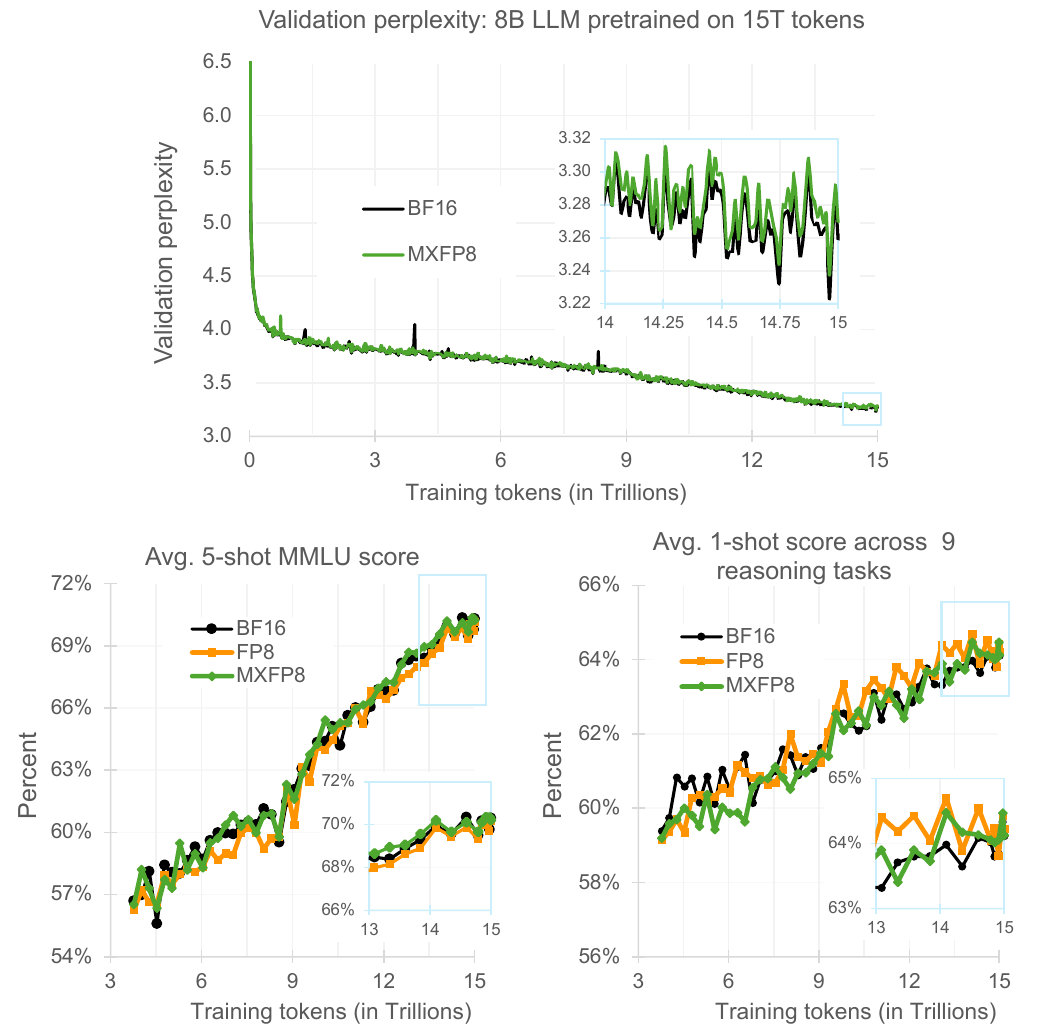}
    \caption{Pre-training a 8B LLM on 15T tokens. \textbf{Top:} Training behavior of \texttt{BF16} vs \texttt{MXFP8}. \textbf{Bottom:} Comparing \texttt{BF16}, \texttt{FP8}  and \texttt{MXFP8}'s downstream task scores on MMLU and a set of 9 reasoning tasks. \texttt{MXFP8} numerics use our proposed rounding method and \texttt{E4M3} for all quantized tensors.} 
    \label{fig:8b_15t_training}
\end{figure*}

\section{Pre-training with MXFP8}
\label{sec:training}

In this section we review \texttt{MXFP8} training results, training recipe, and the algorithm for converting to \texttt{MXFP8} used for training. Ablation studies explain the recipe and conversion algorithm choices.

\subsection{Training Results}
\label{sec:results_mxfp8}

To establish the equivalence of \texttt{MXFP8} and \texttt{BF16} for model accuracy we pretrain an 8B parameter model on 15T tokens. We observe:

\begin{enumerate}[label={--}]

\item Validation perplexity when using \texttt{MXFP8} matches that of \texttt{BF16} (top plot in Figure~\ref{fig:8b_15t_training}). There is less than 0.50\% difference between \texttt{MXFP8} and  \texttt{BF16} validation perplexity values throughout the pre-training run.

\item Downstream task evaluation scores match between \texttt{MXFP8} and \texttt{BF16} sessions (bottom plots in Figure~\ref{fig:8b_15t_training}).
\end{enumerate}

The same equivalence is also observed for smaller models or smaller datasets, making \texttt{MXFP8} a preferred option to efficiently train LLMs. The 8B parameter model~\cite{parmar2024nemotron415btechnicalreport} consists of 32 transformer blocks, 32 attention heads. Hidden size is 4096, GQA group size is 8, KV-channels count is 128, sequence length during pre-training is 8192. Initial learning rate is 6e-4 that cosine decays to 6e-6. A phased data-blending approach is used to train the model: in the first phase, a data mixture that promotes diversity in data is used and in the second phase high-quality datasets (e.g., Wikipedia) are used. We switch to the second phase at the 60\% point of training. This blend style has also been used in other large scale pre-training setups~\cite{nvidia2025nemotronhfamilyaccurateefficient}.

Training was carried out using Megatron-LM software~\cite{shoeybi2020megatronlmtrainingmultibillionparameter} on 3072 Hopper GPUs using batch size 768. \texttt{MX} operations were simulated by converting BF16 inputs to \texttt{MXFP8} and back to \texttt{BF16} prior to GEMM operations.

Evaluation scores are measured on two sets of downstream tasks: (1) 5-shot score on MMLU~\cite{hendrycks2021measuringmassivemultitasklanguage} and (2) Averaged 1-shot score across 9 general reasoning benchmarks: ARC-Challenge and ARC-Easy~\cite{allenai:arc}, Race~\cite{lai2017large}, PIQA~\cite{bisk2019piqareasoningphysicalcommonsense}, Winogrande~\cite{sakaguchi2019winograndeadversarialwinogradschema}, Hellaswag~\cite{zellers2019hellaswagmachinereallyfinish}, OpenBookQA~\cite{mihaylov2018suitarmorconductelectricity}, Social IQA~\cite{sap2019socialiqacommonsensereasoningsocial} and Commonsense QA~\cite{talmor-etal-2019-commonsenseqa}. 

In summary, we find \texttt{MXFP8} maintains accuracy compared to \texttt{BF16} or \texttt{FP8} pre-trained models. On Blackwell GPU-based systems, \texttt{MXFP8} has 2$\times$ higher throughput than \texttt{BF16} making end-to-end \texttt{MXFP8} pre-training faster than \texttt{BF16} pre-training. We also find the \texttt{MXFP8} recipe to be simpler to use when compared to \texttt{FP8} (all layers can be quantized and scaling is handled in the hardware) while allowing for equal or better throughput.

\subsection{MXFP8 Training Recipe}
\label{sec:dtype_tensors}

The recipe is defined by two choices: which tensors to quantize to \texttt{MXFP8} and which \texttt{FP8} binary encoding to use for different tensors.

\begin{figure*}[htb] 
    \centering
    \includegraphics[width=.875\textwidth]{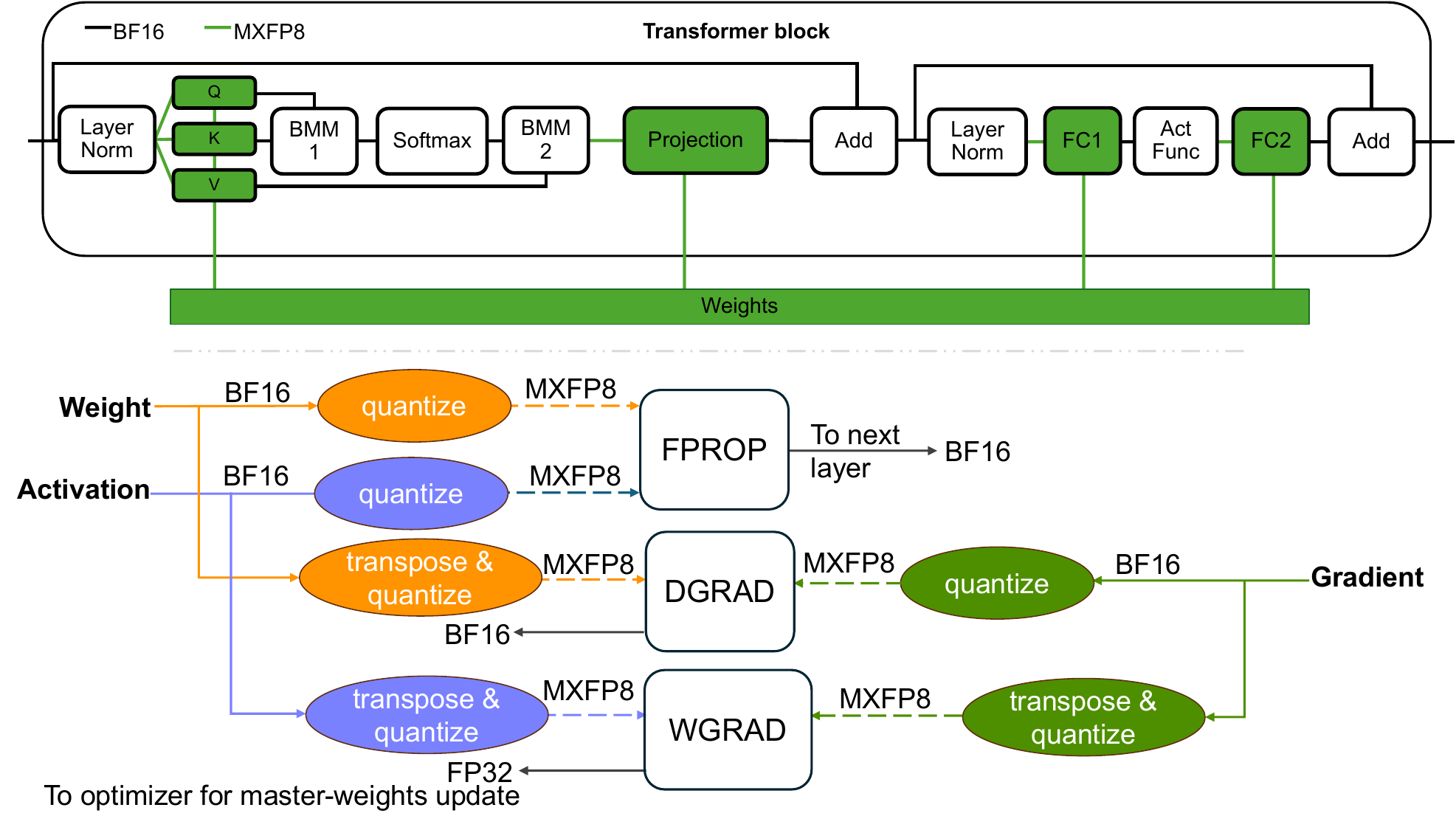}
    \caption{\textbf{Top:} Transformer layers quantized to \texttt{MXFP8} inside a single transformer block. \textbf{Bottom:} Training workflow for a single layer during forward-pass (\texttt{FPROP}), error-gradient pass (\texttt{DGRAD}) and weight-gradient pass (\texttt{WGRAD}).}
    \label{fig:quant_transformer_block_combined}
\end{figure*}

\subsubsection{Tensors to Quantize}
We execute GEMMs with weights (i.e. \texttt{QKV} and \texttt{Proj} in attention and \texttt{FFN}, as shown in Figure~\ref{fig:quant_transformer_block_combined}) in \texttt{MXFP8}. Therefore, activation, weight, as well as activation gradient input tensors for these GEMMs are converted to \texttt{MXFP8}. The Batch-Matrix Multiplications (\texttt{BMM1}, the query-key dot product and \texttt{BMM2}, the attention score-value product) in the self-attention layer, along with operations like \texttt{Softmax}, \texttt{Act-func} and \texttt{residual-add} are in high-precision. The input embedding layer and the final output-projection layer are also in \texttt{BF16} or \texttt{FP16}. We leave it to future work to explore reducing the precision of these operations.

The aforementioned tensors are quantized to \texttt{MXFP8} for \underline{all} transformer blocks of a model, in contrast to per-tensor~\cite{nvidia2025nemotronhfamilyaccurateefficient} or per-row quantization~\cite{grattafiori2024llama3herdmodels} for 
\texttt{FP8}, where some transformer layers were left in \texttt{BF16}. As a result, \texttt{MXFP8} due to its fine-granularity scaling factors enables accelerating more of an LLM's layers.

During training, with \texttt{MXFP} quantization, the training framework keeps two copies of each tensor, each copy is quantized along the axes of dot-product reduction (row and column).
Figure~\ref{fig:quant_transformer_block_combined} shows how each tensor is used in forward (\texttt{FPROP}), weight-gradient (\texttt{WGRAD}) and activation-gradient (\texttt{DGRAD}) computation during the training loop. Since each tensor is used in non-transposed and transposed form, 
quantization needs to occur along two separate axes (row and column).

\begin{figure*}[htbp]
  \centering
  \begin{subfigure}{0.475\linewidth}
    \includegraphics[width=\linewidth]{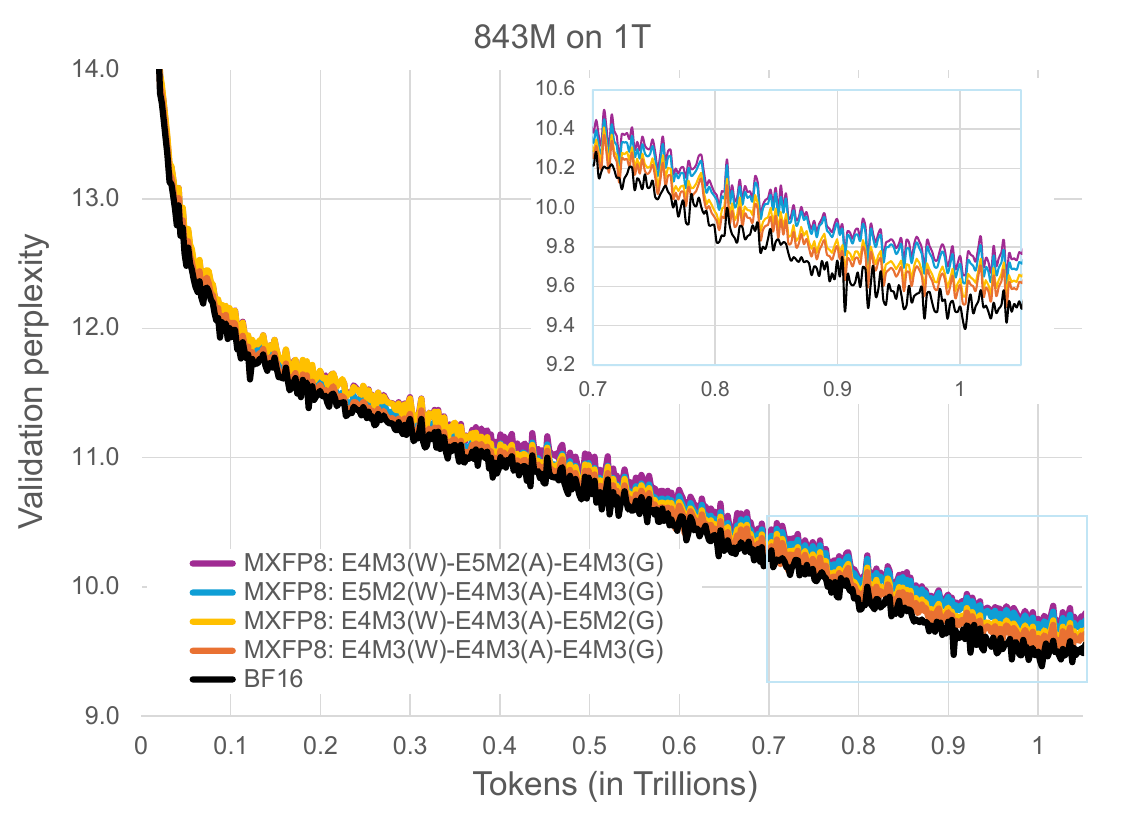}
    \caption{Validation perplexity curves for 843M parameter model trained on 1T tokens.}
    \label{fig:843m_1t_shmoo}
  \end{subfigure}
  \hfill
  \begin{subfigure}{0.475\linewidth}
    \includegraphics[width=\linewidth]{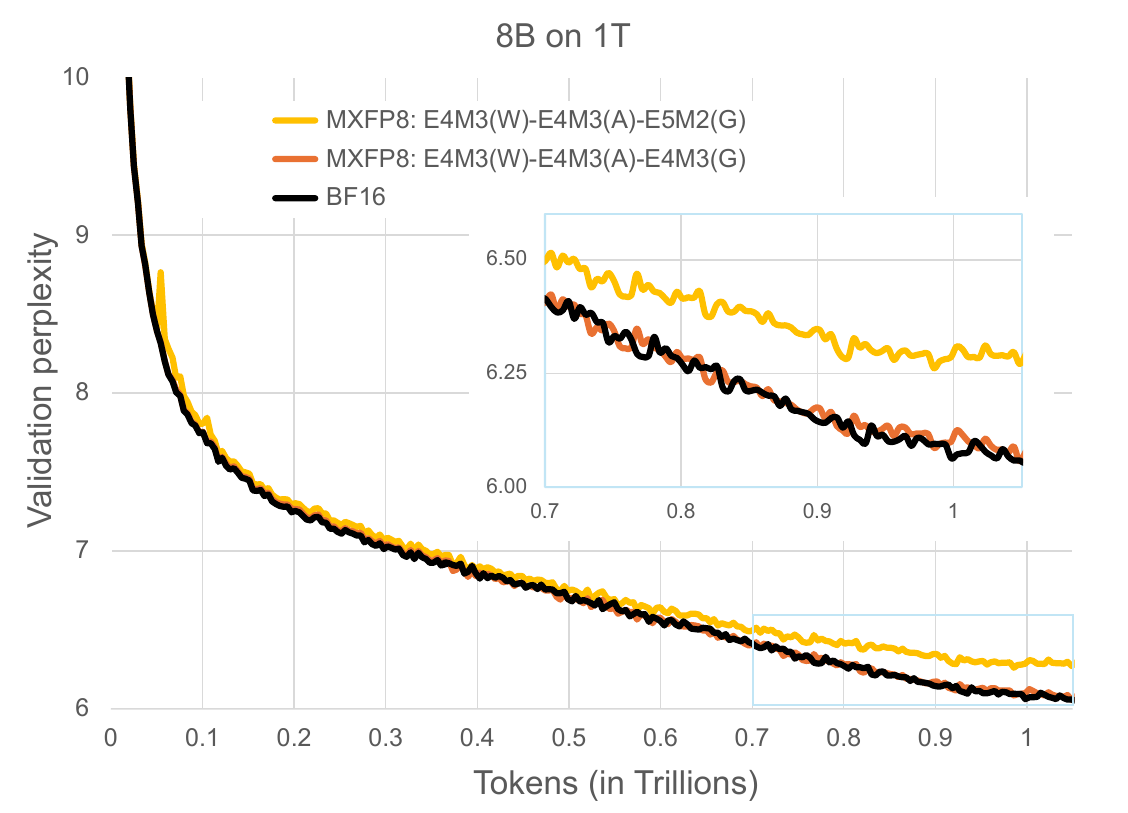}
    \caption{Validation perplexity curves for 8B parameter model trained on 1T tokens.}
    \label{fig:8b_1t}
  \end{subfigure}
  
  \caption{Pre-training loss curves comparing \texttt{E4M3} and \texttt{E5M2} 
   when used across different tensor types: weights (W), activation (A) and gradients (G). 
   The inset shows a zoomed-in view of the loss at the end of training.}
  \label{fig:mxfp8_dtype}
\end{figure*}

\subsubsection{FP8 Encoding Choice}
Our \texttt{MXFP8} recipe uses \texttt{E4M3} encoding for all tensor types - weights, activations, and activation gradients. This is in contrast to coarser-scaled \texttt{FP8} training~\cite{nvidia2025nemotronhfamilyaccurateefficient, micikevicius2022fp8formatsdeeplearning, noune20228bitnumericalformatsdeep, deepseekai2025deepseekv3technicalreport}, which used \texttt{E4M3} for weights and activations but \texttt{E5M2} for gradients. With fine-grained scaling of \texttt{MXFP8}, the dynamic range requirement at a 32-element block size is sufficiently captured by 17.8 binades of \texttt{E4M3} type. Once the range requirements are met, precision (or sampling) becomes important. We demonstrate this recipe choice with an empirical study on an 843M and 8B parameter models. Figure~\ref{fig:843m_1t_shmoo} shows that \texttt{E4M3} is the preferred choice for weights and activations, as using \texttt{E5M2} for these tensor types (the blue and purple curves, respectively) results in a worse perplexity. While either \texttt{E5M2} or \texttt{E4M3} encodings are comparably good for the 843M parameter model, in Figure~\ref{fig:8b_1t} we see that the larger 8B parameter model requires the higher precision of \texttt{E4M3} for activation gradients (orange curve), as \texttt{E5M2} gradients (yellow curve) lead to a substantial degradation in perplexity.

\subsection{Conversion from FP32 to MXFP8}
\label{sec:conversion}

Often the original high precision (for example, \texttt{FP32} or \texttt{BF16}) values within a block are  outside the representable range for the target \texttt{MX} format, both underflowing the minimum and overflowing the maximum representable magnitude. 
To address this, prior to conversion all values are multiplied with a scale factor so that no value overflows the \texttt{MX} format range, while minimizing flushes to 0. 
The shared per-block scale factor, \verb+X+, is computed based on the maximum absolute value (\verb+amax+) among the 32 high-precision input values, i.e. \verb+amax+ = \(\max\|V_{i}\|; 1 \leq i \leq 32\). 
The goal is to map this \verb+amax+ in the input-format to be the largest representable value in the desired \texttt{MX}-format. 
Once \verb+X+ is computed, $V_i$ is divided by $X$ and the resulting value is quantized to a \texttt{FP8}-representable number using Round-to-nearest-ties-to-even (RN) rounding. Furthermore, saturating conversion mode is used (as described in OCP \texttt{FP8} specification~\cite{ocpv1spec}) i.e. magnitudes exceeding \texttt{FP8}-maximum are clamped to respective maximum representative magnitude in quantized format.
Special care is taken if some or all elements in the input are \verb+Infinity+ and/or not-a-number (\verb+NaN+).

\begin{algorithm}[thb]
\caption{Our proposal for computing scale factor, X (simplified method)}\label{alg:scale}
\begin{algorithmic}
\State \(X_{\text{float}} \gets \) \verb+amax+/\verb+destmax+  \Comment{\texttt{destmax} is max representable in \texttt{MX}-format}
\State \(\text{expX}_{\text{float}} \gets \log_2(X_{\text{float}})\) \Comment{extract the exponent of X (de-biased form)}
\State \(\text{expX}_{\text{int}} \gets \text{ceil}(\text{expX}_{\text{float}}) \) \Comment{\textbf{round-up}}
\State \(X \gets \text{clamp}(\text{expX}_{\text{int}}, -127, 127) \) \Comment{clamp to min/max \texttt{E8M0} representable}
\State \(X \gets X + 127 \) \Comment{add bias}
\State return \(2^X\) \Comment{store X}
\end{algorithmic}
\end{algorithm}

Algorithm~\ref{alg:scale} outlines the scale-factor computation used by training sessions in Section~\ref{sec:results_mxfp8}. Step 3 is the key part, \textit{rounding-up the scale factor towards positive infinity}, when converting X to a power-of-2 value. This is different from the scheme suggested by OCP v1 specification, which effectively rounds down the scale factor. Since during quantization to \texttt{MXFP}, high precision values are divided by X, rounding up X ensures that after scaling no $V / X$ value exceeds the maximum magnitude representable in the destination format. Conversely, rounding down, as in OCP v1 suggestion, will lead to scaled values overflowing the representable range, introducing more quantization noise. 

\begin{figure*}[htbp]
  \centering
  \begin{subfigure}{0.475\linewidth}
    \includegraphics[width=\linewidth]{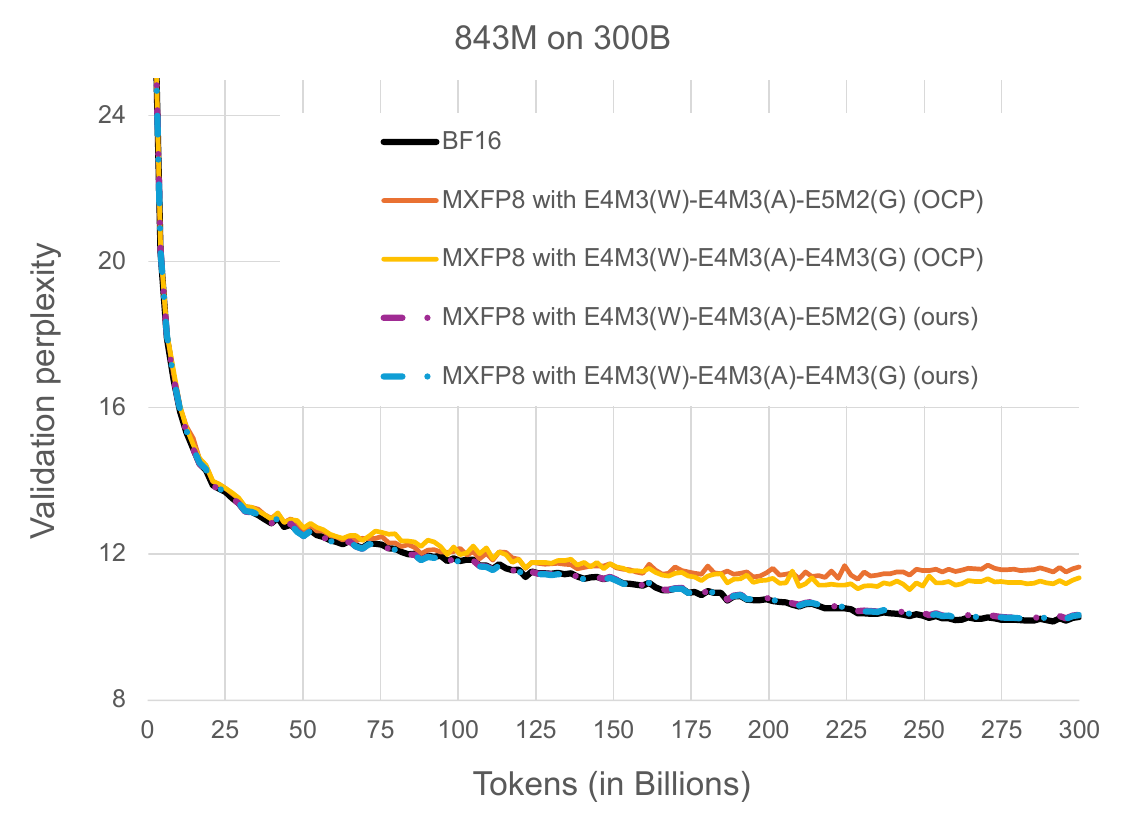}
    \caption{Validation perplexity curves for 843M parameter model trained on 300B tokens.}
    \label{fig:843m_300b}
  \end{subfigure}
  \hfill
  \begin{subfigure}{0.475\linewidth}
    \includegraphics[width=\linewidth]{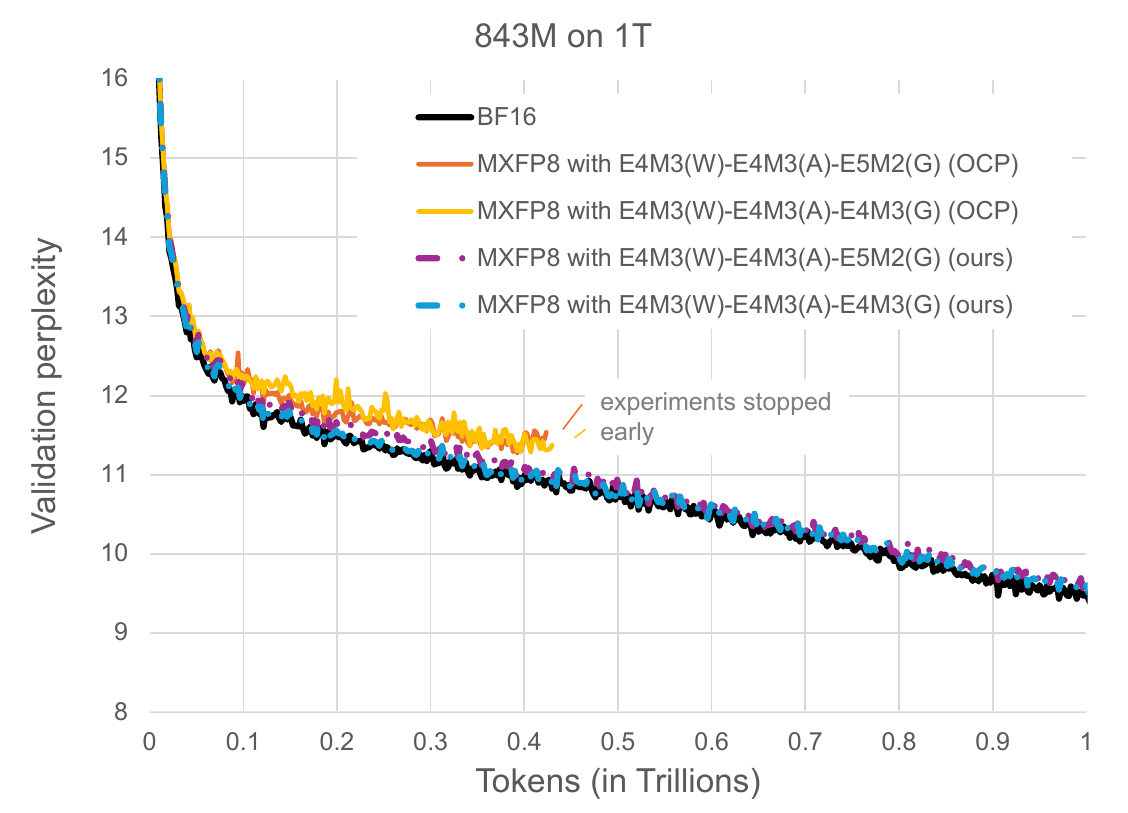}
    \caption{Validation perplexity curves for 843M parameter model trained on 1T tokens.}
    \label{fig:843m_1t}
  \end{subfigure}
  \caption{Comparing scale factor rounding modes suggested in the OCP v1.0 specification and our work. \texttt{E4M3} and \texttt{E5M2}  are \texttt{MXFP8} formats; notation \texttt{E4M3}(W) denotes \texttt{E4M3} data type for Weights. Similarly, \texttt{E4M3}(A) and \texttt{E4M3} (G) refer to activations and gradients, respectively.}
  \label{fig:ocp_vs_ours}
\end{figure*}

We hypothesize this aspect 
leading to \texttt{MXFP8} training issues when using OCP v1 conversion scheme as shown in Figure~\ref{fig:ocp_vs_ours}. As can be seen in the figure, these issues are eliminated when using our proposal in Algorithm\ref{alg:scale}. A more detailed description of conversion workflow is provided in Appendix Sec.~\ref{App_e8m0_rounding}.

\section{Related work}
\label{sec:related_work}
Low-precision training and inference is a widely studied topic in deep learning literature. While significant progress has been made
in low-precision inference~\cite{xiao2024smoothquantaccurateefficientposttraining, lin2025qservew4a8kv4quantizationcodesign, frantar2023gptqaccurateposttrainingquantization, lin2024awqactivationawareweightquantization}, there are relatively fewer studies demonstrating low-precision techniques for pre-training LLMs, especially large-scale pre-training on a large token horizon. Our work primarily focuses on 8-bit pre-training and prior work on related low-precision LLM pre-training can be grouped into the following two categories: 

\begin{enumerate}[label={--}]

\item \textbf{LLM pre-training using FP8 formats:}~\cite{micikevicius2022fp8formatsdeeplearning} proposes an \texttt{FP8} binary interchange format, consisting of \texttt{E4M3} and \texttt{E5M2} encodings, and a per-tensor scaling approach — an entire tensor is scaled to capture the maximum number of tensor values in the representable range in \texttt{FP8}.~\cite{fishman2025scalingfp8trainingtrilliontoken} discusses \texttt{FP8} pre-training challenges and proposes model-level modifications to train a 7B parameter model.
Recently, per-tensor scaled \texttt{FP8} was used to train the Nemotron-H family of LLMs~\cite{nvidia2025nemotronhfamilyaccurateefficient} and the Llama-4 family of models also used \texttt{FP8} ~\cite{meta_llama4_blog}. Both of these approaches mention the need to keep a few transformer blocks in high-precision. Instead of per-tensor scaling, DeepSeek-V3 family of models~\cite{deepseekai2025deepseekv3technicalreport} use block-scaled \texttt{FP8}, this helps to better capture outliers and minimize quantization errors. With block-scaling setup, certain tensors require vector 1x128 scaling and certain tensors require per-block (e.g. 128x128) software scaling. Native support for \texttt{MXFP8} simplifies this — fine-grained scaling provides better numerical robustness and hardware support for scaling avoids any tradeoff between smaller block sizes and hardware speed.

\item \textbf{Pre-training using MXFP formats:}~\cite{rouhani2023microscalingdataformatsdeep} presents empirical data on pre-training models with \texttt{MXFP} formats. They show cast-only inference results for \texttt{MXFP8} and pre-training results for \texttt{MXFP6} and \texttt{MXFP4}-weights.~\cite{tseng2025trainingllmsmxfp4} studies \texttt{MXFP4} backward-pass quantization and~\cite{wang2025optimizinglargelanguagemodel} investigates \texttt{MXFP4} weight quantization on relatively small token horizon.

\end{enumerate}

\section{Conclusions and future work}
\label{sec:conclusions}

We present studies on several aspects of \texttt{MXFP8}-formats and their usage in pre-training models, e.g. scale computation, data format of specific tensors, numerical aspects of quantization, model layers that could be quantized for maximizing training throughput. 
We discuss recipes for pre-training LLMs with \texttt{MXFP8}-formats.
The \texttt{MXFP8} pre-training recipe is implemented in Transformer Engine~\cite{te} and the conversion numerics in cuDNN and cuBLAS libraries.

As part of our future work, we plan to extend our study to post-training phases (e.g. phases that perform reinforcement learning (RL)-based policy optimizations and their variants~\cite{shao2024deepseekmathpushinglimitsmathematical} as well as supervised fine-tuning stages). Additionally, from a hardware systems performance perspective, since tensors have to be quantized along two separate axes with \texttt{MX}-formats, the training framework needs to store two copies for each tensor (we discussed this in section~\ref{sec:dtype_tensors}). We plan to study scaling methods that would lower the storage overhead.

\section*{Acknowledgments}

We thank Sweta Priyadarshi, Mikail Khona and Ben Lanir for helping with the draft revisions as well as Mohammad Shoeybi, Carlo del Mundo, Michael Andersch, Eric Chung and Bryan Catanzaro for valuable feedback and discussions.

\newpage
\bibliography{citations}

\begin{thebibliography}{35}
\providecommand{\natexlab}[1]{#1}
\providecommand{\url}[1]{\texttt{#1}}
\expandafter\ifx\csname urlstyle\endcsname\relax
  \providecommand{\doi}[1]{doi: #1}\else
  \providecommand{\doi}{doi: \begingroup \urlstyle{rm}\Url}\fi

\bibitem[Rouhani et~al.(2023{\natexlab{a}})Rouhani, Garegrat, Savell, More, Han, Zhao, Hall, Klar, Chung, Yu, Schulte, Wittig, Bratt, Stephens, Milanovic, Brothers, Dubey, Cornea, Heinecke, Rodriguez, Langhammer, Deng, Naumov, Micikevicius, Siu, and Verrilli]{ocp}
Bita~Darvish Rouhani, Nitin Garegrat, Tom Savell, Ankit More, Kyung-Nam Han, Ritchie Zhao, Mathew Hall, Jasmine Klar, Eric Chung, Yuan Yu, Michael Schulte, Ralph Wittig, Ian Bratt, Nigel Stephens, Jelena Milanovic, John Brothers, Pradeep Dubey, Marius Cornea, Alexander Heinecke, Andres Rodriguez, Martin Langhammer, Summer Deng, Maxim Naumov, Paulius Micikevicius, Michael Siu, and Colin Verrilli.
\newblock Ocp microscaling (mx) specification.
\newblock \emph{Open Compute Project}, 2023{\natexlab{a}}.

\bibitem[bla()]{blackwell}
Nvidia {B}lackwell {A}rchitecture {T}echnical {B}rief.
\newblock URL \url{https://resources.nvidia.com/en-us-blackwell-architecture}.

\bibitem[Rouhani et~al.(2023{\natexlab{b}})Rouhani, Zhao, More, Hall, Khodamoradi, Deng, Choudhary, Cornea, Dellinger, Denolf, Dusan, Elango, Golub, Heinecke, James-Roxby, Jani, Kolhe, Langhammer, Li, Melnick, Mesmakhosroshahi, Rodriguez, Schulte, Shafipour, Shao, Siu, Dubey, Micikevicius, Naumov, Verrilli, Wittig, Burger, and Chung]{rouhani2023microscalingdataformatsdeep}
Bita~Darvish Rouhani, Ritchie Zhao, Ankit More, Mathew Hall, Alireza Khodamoradi, Summer Deng, Dhruv Choudhary, Marius Cornea, Eric Dellinger, Kristof Denolf, Stosic Dusan, Venmugil Elango, Maximilian Golub, Alexander Heinecke, Phil James-Roxby, Dharmesh Jani, Gaurav Kolhe, Martin Langhammer, Ada Li, Levi Melnick, Maral Mesmakhosroshahi, Andres Rodriguez, Michael Schulte, Rasoul Shafipour, Lei Shao, Michael Siu, Pradeep Dubey, Paulius Micikevicius, Maxim Naumov, Colin Verrilli, Ralph Wittig, Doug Burger, and Eric Chung.
\newblock Microscaling data formats for deep learning, 2023{\natexlab{b}}.
\newblock URL \url{https://arxiv.org/abs/2310.10537}.

\bibitem[Rouhani et~al.(2023{\natexlab{c}})Rouhani, Zhao, Elango, Shafipour, Hall, Mesmakhosroshahi, More, Melnick, Golub, Varatkar, Shao, Kolhe, Melts, Klar, L'Heureux, Perry, Burger, Chung, Deng, Naghshineh, Park, and Naumov]{rouhani2023sharedmicroexponentslittleshifting}
Bita Rouhani, Ritchie Zhao, Venmugil Elango, Rasoul Shafipour, Mathew Hall, Maral Mesmakhosroshahi, Ankit More, Levi Melnick, Maximilian Golub, Girish Varatkar, Lei Shao, Gaurav Kolhe, Dimitry Melts, Jasmine Klar, Renee L'Heureux, Matt Perry, Doug Burger, Eric Chung, Zhaoxia Deng, Sam Naghshineh, Jongsoo Park, and Maxim Naumov.
\newblock With shared microexponents, a little shifting goes a long way, 2023{\natexlab{c}}.
\newblock URL \url{https://arxiv.org/abs/2302.08007}.

\bibitem[Dai et~al.(2021)Dai, Venkatesan, Ren, Zimmer, Dally, and Khailany]{dai2021vsquantpervectorscaledquantization}
Steve Dai, Rangharajan Venkatesan, Haoxing Ren, Brian Zimmer, William~J. Dally, and Brucek Khailany.
\newblock Vs-quant: Per-vector scaled quantization for accurate low-precision neural network inference, 2021.
\newblock URL \url{https://arxiv.org/abs/2102.04503}.

\bibitem[iee(2008)]{ieeefp}
Ieee standard for floating-point arithmetic.
\newblock \emph{IEEE Std 754-2008}, pages 1--70, 2008.
\newblock \doi{10.1109/IEEESTD.2008.4610935}.

\bibitem[Micikevicius et~al.(2022)Micikevicius, Stosic, Burgess, Cornea, Dubey, Grisenthwaite, Ha, Heinecke, Judd, Kamalu, Mellempudi, Oberman, Shoeybi, Siu, and Wu]{micikevicius2022fp8formatsdeeplearning}
Paulius Micikevicius, Dusan Stosic, Neil Burgess, Marius Cornea, Pradeep Dubey, Richard Grisenthwaite, Sangwon Ha, Alexander Heinecke, Patrick Judd, John Kamalu, Naveen Mellempudi, Stuart Oberman, Mohammad Shoeybi, Michael Siu, and Hao Wu.
\newblock Fp8 formats for deep learning, 2022.
\newblock URL \url{https://arxiv.org/abs/2209.05433}.

\bibitem[Parmar et~al.(2024)Parmar, Prabhumoye, Jennings, Patwary, Subramanian, Su, Zhu, Narayanan, Jhunjhunwala, Dattagupta, Jawa, Liu, Mahabaleshwarkar, Nitski, Brundyn, Maki, Martinez, You, Kamalu, LeGresley, Fridman, Casper, Aithal, Kuchaiev, Shoeybi, Cohen, and Catanzaro]{parmar2024nemotron415btechnicalreport}
Jupinder Parmar, Shrimai Prabhumoye, Joseph Jennings, Mostofa Patwary, Sandeep Subramanian, Dan Su, Chen Zhu, Deepak Narayanan, Aastha Jhunjhunwala, Ayush Dattagupta, Vibhu Jawa, Jiwei Liu, Ameya Mahabaleshwarkar, Osvald Nitski, Annika Brundyn, James Maki, Miguel Martinez, Jiaxuan You, John Kamalu, Patrick LeGresley, Denys Fridman, Jared Casper, Ashwath Aithal, Oleksii Kuchaiev, Mohammad Shoeybi, Jonathan Cohen, and Bryan Catanzaro.
\newblock Nemotron-4 15b technical report, 2024.
\newblock URL \url{https://arxiv.org/abs/2402.16819}.

\bibitem[NVIDIA et~al.(2025)NVIDIA, :, Blakeman, Basant, Khattar, Renduchintala, Bercovich, Ficek, Bjorlin, Taghibakhshi, Deshmukh, Mahabaleshwarkar, Tao, Shors, Aithal, Poojary, Dattagupta, Buddharaju, Chen, Ginsburg, Wang, Norick, Butterfield, Catanzaro, del Mundo, Dong, Harvey, Parisien, Su, Korzekwa, Yin, Gitman, Mosallanezhad, Narayanan, Fridman, Rekesh, Ma, Pykhtar, Ahn, Riach, Stosic, Long, Segal, Evans, Chung, Galinkin, Bakhturina, Dobrowolska, Jia, Liu, Prasad, Shen, Liu, Chen, Qian, Ngo, Liu, Li, Gitman, Karmanov, Moshkov, Golan, Kautz, Scowcroft, Casper, Seppanen, Lu, Sewall, Zeng, You, Zhang, Zhang, Huang, Xue, Huang, Conway, Kamalu, Barker, Cohen, Jennings, Parmar, Sapra, Briski, Chumachenko, Luna, Santhanam, Kong, Sivamani, Pawelec, Anik, Li, McAfee, Derczynski, Pavao, Vega, Voegtle, Bala, de~Melo, Sreedhar, Chochowski, Kliegl, Stepniewska-Dziubinska, Le, Novikov, Samadi, Andersch, Evans, Martinez, Chrzanowski, Ranzinger, Blaz, Smelyanskiy, Fawzy, Shoeybi, Patwary, Lee, Tajbakhsh, Xu, Rybakov,
  Kuchaiev, Delalleau, Nitski, Chadha, Shamis, Micikevicius, Molchanov, Dykas, Fischer, Aquilanti, Bialecki, Varshney, Gundecha, Tredak, Karimi, Kandu, El-Yaniv, Joshi, Waleffe, Zhang, Kavanaugh, Jain, Kriman, Lym, Satheesh, Muralidharan, Narenthiran, Anandaraj, Bak, Kashirsky, Han, Acharya, Ghosh, Sreenivas, Clay, Thomas, Prabhumoye, Pachori, Toshniwal, Prayaga, Jain, Das, Kierat, Majumdar, Han, Singhal, Niverty, Alborghetti, Panguluri, Bhendigeri, Akter, Migacz, Shiri, Kong, Roman, Ronen, Saar, Konuk, Rintamaki, Poon, De, Noroozi, Singh, Korthikanti, Kurin, Ahmad, Du, Ping, Dai, Byeon, Ren, Xu, Choi, Zhang, Lin, Suhara, Yu, Li, Li, Zhu, Yang, and Chen]{nvidia2025nemotronhfamilyaccurateefficient}
NVIDIA, :, Aaron Blakeman, Aarti Basant, Abhinav Khattar, Adithya Renduchintala, Akhiad Bercovich, Aleksander Ficek, Alexis Bjorlin, Ali Taghibakhshi, Amala~Sanjay Deshmukh, Ameya~Sunil Mahabaleshwarkar, Andrew Tao, Anna Shors, Ashwath Aithal, Ashwin Poojary, Ayush Dattagupta, Balaram Buddharaju, Bobby Chen, Boris Ginsburg, Boxin Wang, Brandon Norick, Brian Butterfield, Bryan Catanzaro, Carlo del Mundo, Chengyu Dong, Christine Harvey, Christopher Parisien, Dan Su, Daniel Korzekwa, Danny Yin, Daria Gitman, David Mosallanezhad, Deepak Narayanan, Denys Fridman, Dima Rekesh, Ding Ma, Dmytro Pykhtar, Dong Ahn, Duncan Riach, Dusan Stosic, Eileen Long, Elad Segal, Ellie Evans, Eric Chung, Erick Galinkin, Evelina Bakhturina, Ewa Dobrowolska, Fei Jia, Fuxiao Liu, Gargi Prasad, Gerald Shen, Guilin Liu, Guo Chen, Haifeng Qian, Helen Ngo, Hongbin Liu, Hui Li, Igor Gitman, Ilia Karmanov, Ivan Moshkov, Izik Golan, Jan Kautz, Jane~Polak Scowcroft, Jared Casper, Jarno Seppanen, Jason Lu, Jason Sewall, Jiaqi Zeng, Jiaxuan
  You, Jimmy Zhang, Jing Zhang, Jining Huang, Jinze Xue, Jocelyn Huang, Joey Conway, John Kamalu, Jon Barker, Jonathan Cohen, Joseph Jennings, Jupinder Parmar, Karan Sapra, Kari Briski, Kateryna Chumachenko, Katherine Luna, Keshav Santhanam, Kezhi Kong, Kirthi Sivamani, Krzysztof Pawelec, Kumar Anik, Kunlun Li, Lawrence McAfee, Leon Derczynski, Lindsey Pavao, Luis Vega, Lukas Voegtle, Maciej Bala, Maer~Rodrigues de~Melo, Makesh~Narsimhan Sreedhar, Marcin Chochowski, Markus Kliegl, Marta Stepniewska-Dziubinska, Matthieu Le, Matvei Novikov, Mehrzad Samadi, Michael Andersch, Michael Evans, Miguel Martinez, Mike Chrzanowski, Mike Ranzinger, Mikolaj Blaz, Misha Smelyanskiy, Mohamed Fawzy, Mohammad Shoeybi, Mostofa Patwary, Nayeon Lee, Nima Tajbakhsh, Ning Xu, Oleg Rybakov, Oleksii Kuchaiev, Olivier Delalleau, Osvald Nitski, Parth Chadha, Pasha Shamis, Paulius Micikevicius, Pavlo Molchanov, Peter Dykas, Philipp Fischer, Pierre-Yves Aquilanti, Piotr Bialecki, Prasoon Varshney, Pritam Gundecha, Przemek Tredak, Rabeeh
  Karimi, Rahul Kandu, Ran El-Yaniv, Raviraj Joshi, Roger Waleffe, Ruoxi Zhang, Sabrina Kavanaugh, Sahil Jain, Samuel Kriman, Sangkug Lym, Sanjeev Satheesh, Saurav Muralidharan, Sean Narenthiran, Selvaraj Anandaraj, Seonmyeong Bak, Sergey Kashirsky, Seungju Han, Shantanu Acharya, Shaona Ghosh, Sharath~Turuvekere Sreenivas, Sharon Clay, Shelby Thomas, Shrimai Prabhumoye, Shubham Pachori, Shubham Toshniwal, Shyamala Prayaga, Siddhartha Jain, Sirshak Das, Slawek Kierat, Somshubra Majumdar, Song Han, Soumye Singhal, Sriharsha Niverty, Stefania Alborghetti, Suseella Panguluri, Swetha Bhendigeri, Syeda~Nahida Akter, Szymon Migacz, Tal Shiri, Terry Kong, Timo Roman, Tomer Ronen, Trisha Saar, Tugrul Konuk, Tuomas Rintamaki, Tyler Poon, Ushnish De, Vahid Noroozi, Varun Singh, Vijay Korthikanti, Vitaly Kurin, Wasi~Uddin Ahmad, Wei Du, Wei Ping, Wenliang Dai, Wonmin Byeon, Xiaowei Ren, Yao Xu, Yejin Choi, Yian Zhang, Ying Lin, Yoshi Suhara, Zhiding Yu, Zhiqi Li, Zhiyu Li, Zhongbo Zhu, Zhuolin Yang, and Zijia Chen.
\newblock Nemotron-h: A family of accurate and efficient hybrid mamba-transformer models, 2025.
\newblock URL \url{https://arxiv.org/abs/2504.03624}.

\bibitem[Shoeybi et~al.(2020)Shoeybi, Patwary, Puri, LeGresley, Casper, and Catanzaro]{shoeybi2020megatronlmtrainingmultibillionparameter}
Mohammad Shoeybi, Mostofa Patwary, Raul Puri, Patrick LeGresley, Jared Casper, and Bryan Catanzaro.
\newblock Megatron-lm: Training multi-billion parameter language models using model parallelism, 2020.
\newblock URL \url{https://arxiv.org/abs/1909.08053}.

\bibitem[Hendrycks et~al.(2021)Hendrycks, Burns, Basart, Zou, Mazeika, Song, and Steinhardt]{hendrycks2021measuringmassivemultitasklanguage}
Dan Hendrycks, Collin Burns, Steven Basart, Andy Zou, Mantas Mazeika, Dawn Song, and Jacob Steinhardt.
\newblock Measuring massive multitask language understanding, 2021.
\newblock URL \url{https://arxiv.org/abs/2009.03300}.

\bibitem[Clark et~al.(2018)Clark, Cowhey, Etzioni, Khot, Sabharwal, Schoenick, and Tafjord]{allenai:arc}
Peter Clark, Isaac Cowhey, Oren Etzioni, Tushar Khot, Ashish Sabharwal, Carissa Schoenick, and Oyvind Tafjord.
\newblock Think you have solved question answering? try arc, the ai2 reasoning challenge.
\newblock \emph{arXiv:1803.05457v1}, 2018.

\bibitem[Lai et~al.(2017)Lai, Xie, Liu, Yang, and Hovy]{lai2017large}
Guokun Lai, Qizhe Xie, Hanxiao Liu, Yiming Yang, and Eduard Hovy.
\newblock Race: Large-scale reading comprehension dataset from examinations.
\newblock \emph{arXiv preprint arXiv:1704.04683}, 2017.

\bibitem[Bisk et~al.(2019)Bisk, Zellers, Bras, Gao, and Choi]{bisk2019piqareasoningphysicalcommonsense}
Yonatan Bisk, Rowan Zellers, Ronan~Le Bras, Jianfeng Gao, and Yejin Choi.
\newblock Piqa: Reasoning about physical commonsense in natural language, 2019.
\newblock URL \url{https://arxiv.org/abs/1911.11641}.

\bibitem[Sakaguchi et~al.(2019)Sakaguchi, Bras, Bhagavatula, and Choi]{sakaguchi2019winograndeadversarialwinogradschema}
Keisuke Sakaguchi, Ronan~Le Bras, Chandra Bhagavatula, and Yejin Choi.
\newblock Winogrande: An adversarial winograd schema challenge at scale, 2019.
\newblock URL \url{https://arxiv.org/abs/1907.10641}.

\bibitem[Zellers et~al.(2019)Zellers, Holtzman, Bisk, Farhadi, and Choi]{zellers2019hellaswagmachinereallyfinish}
Rowan Zellers, Ari Holtzman, Yonatan Bisk, Ali Farhadi, and Yejin Choi.
\newblock Hellaswag: Can a machine really finish your sentence?, 2019.
\newblock URL \url{https://arxiv.org/abs/1905.07830}.

\bibitem[Mihaylov et~al.(2018)Mihaylov, Clark, Khot, and Sabharwal]{mihaylov2018suitarmorconductelectricity}
Todor Mihaylov, Peter Clark, Tushar Khot, and Ashish Sabharwal.
\newblock Can a suit of armor conduct electricity? a new dataset for open book question answering, 2018.
\newblock URL \url{https://arxiv.org/abs/1809.02789}.

\bibitem[Sap et~al.(2019)Sap, Rashkin, Chen, LeBras, and Choi]{sap2019socialiqacommonsensereasoningsocial}
Maarten Sap, Hannah Rashkin, Derek Chen, Ronan LeBras, and Yejin Choi.
\newblock Socialiqa: Commonsense reasoning about social interactions, 2019.
\newblock URL \url{https://arxiv.org/abs/1904.09728}.

\bibitem[Talmor et~al.(2019)Talmor, Herzig, Lourie, and Berant]{talmor-etal-2019-commonsenseqa}
Alon Talmor, Jonathan Herzig, Nicholas Lourie, and Jonathan Berant.
\newblock {C}ommonsense{QA}: A question answering challenge targeting commonsense knowledge.
\newblock In \emph{Proceedings of the 2019 Conference of the North {A}merican Chapter of the Association for Computational Linguistics: Human Language Technologies, Volume 1 (Long and Short Papers)}, pages 4149--4158, Minneapolis, Minnesota, June 2019. Association for Computational Linguistics.
\newblock \doi{10.18653/v1/N19-1421}.
\newblock URL \url{https://aclanthology.org/N19-1421}.

\bibitem[et~al.(2024)]{grattafiori2024llama3herdmodels}
Aaron~Grattafiori et~al.
\newblock The llama 3 herd of models, 2024.
\newblock URL \url{https://arxiv.org/abs/2407.21783}.

\bibitem[Noune et~al.(2022)Noune, Jones, Justus, Masters, and Luschi]{noune20228bitnumericalformatsdeep}
Badreddine Noune, Philip Jones, Daniel Justus, Dominic Masters, and Carlo Luschi.
\newblock 8-bit numerical formats for deep neural networks, 2022.
\newblock URL \url{https://arxiv.org/abs/2206.02915}.

\bibitem[DeepSeek-AI et~al.(2025)DeepSeek-AI, Liu, Feng, Xue, Wang, Wu, Lu, Zhao, Deng, Zhang, Ruan, Dai, Guo, Yang, Chen, Ji, Li, Lin, Dai, Luo, Hao, Chen, Li, Zhang, Bao, Xu, Wang, Zhang, Ding, Xin, Gao, Li, Qu, Cai, Liang, Guo, Ni, Li, Wang, Chen, Chen, Yuan, Qiu, Li, Song, Dong, Hu, Gao, Guan, Huang, Yu, Wang, Zhang, Xu, Xia, Zhao, Wang, Zhang, Li, Wang, Zhang, Zhang, Tang, Li, Tian, Huang, Wang, Zhang, Wang, Zhu, Chen, Du, Chen, Jin, Ge, Zhang, Pan, Wang, Xu, Zhang, Chen, Li, Lu, Zhou, Chen, Wu, Ye, Ye, Ma, Wang, Zhou, Yu, Zhou, Pan, Wang, Yun, Pei, Sun, Xiao, Zeng, Zhao, An, Liu, Liang, Gao, Yu, Zhang, Li, Jin, Wang, Bi, Liu, Wang, Shen, Chen, Zhang, Chen, Nie, Sun, Wang, Cheng, Liu, Xie, Liu, Yu, Song, Shan, Zhou, Yang, Li, Su, Lin, Li, Wang, Wei, Zhu, Zhang, Xu, Xu, Huang, Li, Zhao, Sun, Li, Wang, Yu, Zheng, Zhang, Shi, Xiong, He, Tang, Piao, Wang, Tan, Ma, Liu, Guo, Wu, Ou, Zhu, Wang, Gong, Zou, He, Zha, Xiong, Ma, Yan, Luo, You, Liu, Zhou, Wu, Ren, Ren, Sha, Fu, Xu, Huang, Zhang, Xie, Zhang, Hao,
  Gou, Ma, Yan, Shao, Xu, Wu, Zhang, Li, Gu, Zhu, Liu, Li, Xie, Song, Gao, and Pan]{deepseekai2025deepseekv3technicalreport}
DeepSeek-AI, Aixin Liu, Bei Feng, Bing Xue, Bingxuan Wang, Bochao Wu, Chengda Lu, Chenggang Zhao, Chengqi Deng, Chenyu Zhang, Chong Ruan, Damai Dai, Daya Guo, Dejian Yang, Deli Chen, Dongjie Ji, Erhang Li, Fangyun Lin, Fucong Dai, Fuli Luo, Guangbo Hao, Guanting Chen, Guowei Li, H.~Zhang, Han Bao, Hanwei Xu, Haocheng Wang, Haowei Zhang, Honghui Ding, Huajian Xin, Huazuo Gao, Hui Li, Hui Qu, J.~L. Cai, Jian Liang, Jianzhong Guo, Jiaqi Ni, Jiashi Li, Jiawei Wang, Jin Chen, Jingchang Chen, Jingyang Yuan, Junjie Qiu, Junlong Li, Junxiao Song, Kai Dong, Kai Hu, Kaige Gao, Kang Guan, Kexin Huang, Kuai Yu, Lean Wang, Lecong Zhang, Lei Xu, Leyi Xia, Liang Zhao, Litong Wang, Liyue Zhang, Meng Li, Miaojun Wang, Mingchuan Zhang, Minghua Zhang, Minghui Tang, Mingming Li, Ning Tian, Panpan Huang, Peiyi Wang, Peng Zhang, Qiancheng Wang, Qihao Zhu, Qinyu Chen, Qiushi Du, R.~J. Chen, R.~L. Jin, Ruiqi Ge, Ruisong Zhang, Ruizhe Pan, Runji Wang, Runxin Xu, Ruoyu Zhang, Ruyi Chen, S.~S. Li, Shanghao Lu, Shangyan Zhou, Shanhuang
  Chen, Shaoqing Wu, Shengfeng Ye, Shengfeng Ye, Shirong Ma, Shiyu Wang, Shuang Zhou, Shuiping Yu, Shunfeng Zhou, Shuting Pan, T.~Wang, Tao Yun, Tian Pei, Tianyu Sun, W.~L. Xiao, Wangding Zeng, Wanjia Zhao, Wei An, Wen Liu, Wenfeng Liang, Wenjun Gao, Wenqin Yu, Wentao Zhang, X.~Q. Li, Xiangyue Jin, Xianzu Wang, Xiao Bi, Xiaodong Liu, Xiaohan Wang, Xiaojin Shen, Xiaokang Chen, Xiaokang Zhang, Xiaosha Chen, Xiaotao Nie, Xiaowen Sun, Xiaoxiang Wang, Xin Cheng, Xin Liu, Xin Xie, Xingchao Liu, Xingkai Yu, Xinnan Song, Xinxia Shan, Xinyi Zhou, Xinyu Yang, Xinyuan Li, Xuecheng Su, Xuheng Lin, Y.~K. Li, Y.~Q. Wang, Y.~X. Wei, Y.~X. Zhu, Yang Zhang, Yanhong Xu, Yanhong Xu, Yanping Huang, Yao Li, Yao Zhao, Yaofeng Sun, Yaohui Li, Yaohui Wang, Yi~Yu, Yi~Zheng, Yichao Zhang, Yifan Shi, Yiliang Xiong, Ying He, Ying Tang, Yishi Piao, Yisong Wang, Yixuan Tan, Yiyang Ma, Yiyuan Liu, Yongqiang Guo, Yu~Wu, Yuan Ou, Yuchen Zhu, Yuduan Wang, Yue Gong, Yuheng Zou, Yujia He, Yukun Zha, Yunfan Xiong, Yunxian Ma, Yuting Yan, Yuxiang
  Luo, Yuxiang You, Yuxuan Liu, Yuyang Zhou, Z.~F. Wu, Z.~Z. Ren, Zehui Ren, Zhangli Sha, Zhe Fu, Zhean Xu, Zhen Huang, Zhen Zhang, Zhenda Xie, Zhengyan Zhang, Zhewen Hao, Zhibin Gou, Zhicheng Ma, Zhigang Yan, Zhihong Shao, Zhipeng Xu, Zhiyu Wu, Zhongyu Zhang, Zhuoshu Li, Zihui Gu, Zijia Zhu, Zijun Liu, Zilin Li, Ziwei Xie, Ziyang Song, Ziyi Gao, and Zizheng Pan.
\newblock Deepseek-v3 technical report, 2025.
\newblock URL \url{https://arxiv.org/abs/2412.19437}.

\bibitem[Micikevicius et~al.()Micikevicius, Oberman, Dubey, Cornea, Rodriguez, Bratt, Grisenthwaite, Jouppi, Chou, Huffman, Schulte, Wittig, Jani, and Deng]{ocpv1spec}
Paulius Micikevicius, Stuart Oberman, Pradeep Dubey, Marius Cornea, Andres Rodriguez, Ian Bratt, Richard Grisenthwaite, Norm Jouppi, Chiachen Chou, Amber Huffman, Michael Schulte, Ralph Wittig, Dharmesh Jani, and Summer Deng.
\newblock Ocp 8-bit floating point specification (ofp8).
\newblock URL \url{https://www.opencompute.org/documents/ocp-8-bit-floating-point-specification-ofp8-revision-1-0-2023-12-01-pdf-1}.

\bibitem[Xiao et~al.(2024)Xiao, Lin, Seznec, Wu, Demouth, and Han]{xiao2024smoothquantaccurateefficientposttraining}
Guangxuan Xiao, Ji~Lin, Mickael Seznec, Hao Wu, Julien Demouth, and Song Han.
\newblock Smoothquant: Accurate and efficient post-training quantization for large language models, 2024.
\newblock URL \url{https://arxiv.org/abs/2211.10438}.

\bibitem[Lin et~al.(2025)Lin, Tang, Yang, Zhang, Xiao, Gan, and Han]{lin2025qservew4a8kv4quantizationcodesign}
Yujun Lin, Haotian Tang, Shang Yang, Zhekai Zhang, Guangxuan Xiao, Chuang Gan, and Song Han.
\newblock Qserve: W4a8kv4 quantization and system co-design for efficient llm serving, 2025.
\newblock URL \url{https://arxiv.org/abs/2405.04532}.

\bibitem[Frantar et~al.(2023)Frantar, Ashkboos, Hoefler, and Alistarh]{frantar2023gptqaccurateposttrainingquantization}
Elias Frantar, Saleh Ashkboos, Torsten Hoefler, and Dan Alistarh.
\newblock Gptq: Accurate post-training quantization for generative pre-trained transformers, 2023.
\newblock URL \url{https://arxiv.org/abs/2210.17323}.

\bibitem[Lin et~al.(2024)Lin, Tang, Tang, Yang, Chen, Wang, Xiao, Dang, Gan, and Han]{lin2024awqactivationawareweightquantization}
Ji~Lin, Jiaming Tang, Haotian Tang, Shang Yang, Wei-Ming Chen, Wei-Chen Wang, Guangxuan Xiao, Xingyu Dang, Chuang Gan, and Song Han.
\newblock Awq: Activation-aware weight quantization for llm compression and acceleration, 2024.
\newblock URL \url{https://arxiv.org/abs/2306.00978}.

\bibitem[Fishman et~al.(2025)Fishman, Chmiel, Banner, and Soudry]{fishman2025scalingfp8trainingtrilliontoken}
Maxim Fishman, Brian Chmiel, Ron Banner, and Daniel Soudry.
\newblock Scaling fp8 training to trillion-token llms, 2025.
\newblock URL \url{https://arxiv.org/abs/2409.12517}.

\bibitem[AI(2025)]{meta_llama4_blog}
Meta AI.
\newblock The llama 4 herd: The beginning of a new era of natively multimodal ai innovation.
\newblock \url{https://ai.meta.com/blog/llama-4-multimodal-intelligence/}, April 2025.
\newblock Accessed 12 May 2025.

\bibitem[Tseng et~al.(2025)Tseng, Yu, and Park]{tseng2025trainingllmsmxfp4}
Albert Tseng, Tao Yu, and Youngsuk Park.
\newblock Training llms with mxfp4, 2025.
\newblock URL \url{https://arxiv.org/abs/2502.20586}.

\bibitem[Wang et~al.(2025)Wang, Gong, Liu, Zhao, Yang, Guo, Zha, and Cheng]{wang2025optimizinglargelanguagemodel}
Ruizhe Wang, Yeyun Gong, Xiao Liu, Guoshuai Zhao, Ziyue Yang, Baining Guo, Zhengjun Zha, and Peng Cheng.
\newblock Optimizing large language model training using fp4 quantization, 2025.
\newblock URL \url{https://arxiv.org/abs/2501.17116}.

\bibitem[Nvidia()]{te}
Nvidia.
\newblock Transformer engine.
\newblock \url{https://github.com/NVIDIA/TransformerEngine/}.

\bibitem[Shao et~al.(2024)Shao, Wang, Zhu, Xu, Song, Bi, Zhang, Zhang, Li, Wu, and Guo]{shao2024deepseekmathpushinglimitsmathematical}
Zhihong Shao, Peiyi Wang, Qihao Zhu, Runxin Xu, Junxiao Song, Xiao Bi, Haowei Zhang, Mingchuan Zhang, Y.~K. Li, Y.~Wu, and Daya Guo.
\newblock Deepseekmath: Pushing the limits of mathematical reasoning in open language models, 2024.
\newblock URL \url{https://arxiv.org/abs/2402.03300}.

\bibitem[Wen et~al.(2024)Wen, Li, Wang, Hall, Liang, and Ma]{wen2024understandingwarmupstabledecaylearningrates}
Kaiyue Wen, Zhiyuan Li, Jason Wang, David Hall, Percy Liang, and Tengyu Ma.
\newblock Understanding warmup-stable-decay learning rates: A river valley loss landscape perspective, 2024.
\newblock URL \url{https://arxiv.org/abs/2410.05192}.

\bibitem[Nvidia et~al.(2024)Nvidia, :, Adler, Agarwal, Aithal, Anh, Bhattacharya, Brundyn, Casper, Catanzaro, Clay, Cohen, Das, Dattagupta, Delalleau, Derczynski, Dong, Egert, Evans, Ficek, Fridman, Ghosh, Ginsburg, Gitman, Grzegorzek, Hero, Huang, Jawa, Jennings, Jhunjhunwala, Kamalu, Khan, Kuchaiev, LeGresley, Li, Liu, Liu, Long, Mahabaleshwarkar, Majumdar, Maki, Martinez, de~Melo, Moshkov, Narayanan, Narenthiran, Navarro, Nguyen, Nitski, Noroozi, Nutheti, Parisien, Parmar, Patwary, Pawelec, Ping, Prabhumoye, Roy, Saar, Sabavat, Satheesh, Scowcroft, Sewall, Shamis, Shen, Shoeybi, Sizer, Smelyanskiy, Soares, Sreedhar, Su, Subramanian, Sun, Toshniwal, Wang, Wang, You, Zeng, Zhang, Zhang, Zhang, Zhang, and Zhu]{nvidia2024nemotron4340btechnicalreport}
Nvidia, :, Bo~Adler, Niket Agarwal, Ashwath Aithal, Dong~H. Anh, Pallab Bhattacharya, Annika Brundyn, Jared Casper, Bryan Catanzaro, Sharon Clay, Jonathan Cohen, Sirshak Das, Ayush Dattagupta, Olivier Delalleau, Leon Derczynski, Yi~Dong, Daniel Egert, Ellie Evans, Aleksander Ficek, Denys Fridman, Shaona Ghosh, Boris Ginsburg, Igor Gitman, Tomasz Grzegorzek, Robert Hero, Jining Huang, Vibhu Jawa, Joseph Jennings, Aastha Jhunjhunwala, John Kamalu, Sadaf Khan, Oleksii Kuchaiev, Patrick LeGresley, Hui Li, Jiwei Liu, Zihan Liu, Eileen Long, Ameya~Sunil Mahabaleshwarkar, Somshubra Majumdar, James Maki, Miguel Martinez, Maer~Rodrigues de~Melo, Ivan Moshkov, Deepak Narayanan, Sean Narenthiran, Jesus Navarro, Phong Nguyen, Osvald Nitski, Vahid Noroozi, Guruprasad Nutheti, Christopher Parisien, Jupinder Parmar, Mostofa Patwary, Krzysztof Pawelec, Wei Ping, Shrimai Prabhumoye, Rajarshi Roy, Trisha Saar, Vasanth Rao~Naik Sabavat, Sanjeev Satheesh, Jane~Polak Scowcroft, Jason Sewall, Pavel Shamis, Gerald Shen, Mohammad
  Shoeybi, Dave Sizer, Misha Smelyanskiy, Felipe Soares, Makesh~Narsimhan Sreedhar, Dan Su, Sandeep Subramanian, Shengyang Sun, Shubham Toshniwal, Hao Wang, Zhilin Wang, Jiaxuan You, Jiaqi Zeng, Jimmy Zhang, Jing Zhang, Vivienne Zhang, Yian Zhang, and Chen Zhu.
\newblock Nemotron-4 340b technical report, 2024.
\newblock URL \url{https://arxiv.org/abs/2406.11704}.

\end{thebibliography}

\newpage
\appendix

\section{Appendix}

\subsection{UE8M0 rounding}
\label{App_e8m0_rounding}

Computation described in Algorithm~\ref{alg:scale} is a simplification. The simplification comes from storing the output of log$_2$\verb+(float x)+ in  \texttt{FP32}. log$_2$ function on device internally performs a round-to-nearest of the resulting return value and thus the result of \verb+ceil(+log$_2$\verb+(x))+ can be different if the output of log$_2$ is not stored in a sufficiently large data-type. Hence, for emulation purposes we work directly with the bit representation of the ratio of \verb+amax+ and \verb+destmax+. We, next, describe the computation flow.

\noindent \textbf{Background:} As a reminder, the quantization process from 32 high-precision values, $V_i$, to quantized values, $Q_i$; $1 \leq i \leq 32$, is given by: 
$Q_i$ = \verb+Quantize_to_fp8+$(V_i / X)$.
$X$ is the scale factor; the exponent of $X$ is stored in an 8bit integer container in memory and interpreted as 2 raised to this exponent value by the hardware (after a bias correction). This scale factor \textit{decodes} $Q_i$ back to $V_i$ (with quantization loss). 

Value of scale factor ($X$) = \verb+float_to_8bits(amax/destmax)+, where \verb+amax+ is absolute maximum in input/source block (of 32 elements) and \verb+destmax+ is the largest positive number in the destination (\verb+MX+) number system. \verb+float_to_8bits+ converts a floating-point number to a power-of-two number. 

A float (\texttt{FP32}) number can be represented in IEEE convention as $2^{E}$ $\times$ \verb+1.mantissa+ (normal) or $2^{-126}$ $\times$ \verb+0.mantissa+ (subnormal). \verb+E+ can lie between -127 to 127 (or 0 to 254 with the exponent bias) and can be represented in the 8-bit container for scale factor. -126 (or 1 with the exponent bias) is also representable by 8-bit container. \verb+mantissa+ lies between \verb+[0,1)+. So, the question is: should mantissa bits be rounded-up, rounded-down, round-to-nearest, discarded, etc. to create a power-of-two number? We find round-up to be the best choice for pre-training with \texttt{MX}-formats.

\noindent \textbf{Rounding:} For \verb+float_to_8bits()+, the recommended order of computation is:

\begin{enumerate}[itemsep=-0.1ex]

\item Compute the decoding scale as: \verb+decode_scale+ = \verb+block_amax+/\verb+destmax+

\item if \verb+decode_scale+ is below $2^{-127}$, then set it to $2^{-127}$ (which is the smallest value representable in \verb+UE8M0+)

\item For all other values that are not powers of 2, \textit{round-up} to the closest representable \verb+UE8M0+ value.
    
\end{enumerate}

By construction \verb+amax/destmax+ never exceeds $2^{127}$ (which is the largest value representable in \verb+UE8M0+) with \texttt{FP8}, \texttt{FP6} or \texttt{FP4} formats. The above computations are done in the bit-space in emulation.

\subsection{MX-format conversion}

Section~\ref{sec:mxformat} relies on the standard \texttt{MX}-format conversion algorithm defined in \cite{rouhani2023microscalingdataformatsdeep}, but for completeness we show it here given a shared scaling exponent $X$ as computed in \ref{sec:conversion}.

\noindent \textbf{Quantizing FP32 values to MX type:} After \verb+X+ is computed, $V_i/X$ is computed and the resulting value is quantized to a \texttt{FP8}-representable number (\verb+Quantize_to_fp8()+). Round-to-nearest-ties-to-even (RN) rounding is used during this quantization step. The conversion process is saturating, i.e. if after rounding the resulting value exceeds \texttt{FP8} max or is less than \texttt{FP8} min value, then the result is clamped to respective max or min value.

Quantization operations add computation overhead — Blackwell has hardware support for rounding the scale (using our proposed method) and quantizing values to lower this overhead.

\subsection{Why is special hardware needed for MX-formats?} 

Computing the matrix-product of two tensors involves performing dot-products between sub-vectors of the two tensors. Therefore, scaling factors need to be processed once per group of values that share the scale. Since \texttt{MX}-formats have fine-grained scaling, scale factors are processed once after each block's dot-product is computed, thus, many times per tensor-wide dot-product. This is expensive to do in software, so hardware needs to add support for accelerating tensor operations involving \texttt{MX}-formats (e.g. Blackwell).

\subsection{MXFP8 pre-training for a mixture-of-experts model}\label{app_moe}

\begin{figure*}[hbtp]
    \centering
    \includegraphics[width=0.9\linewidth]{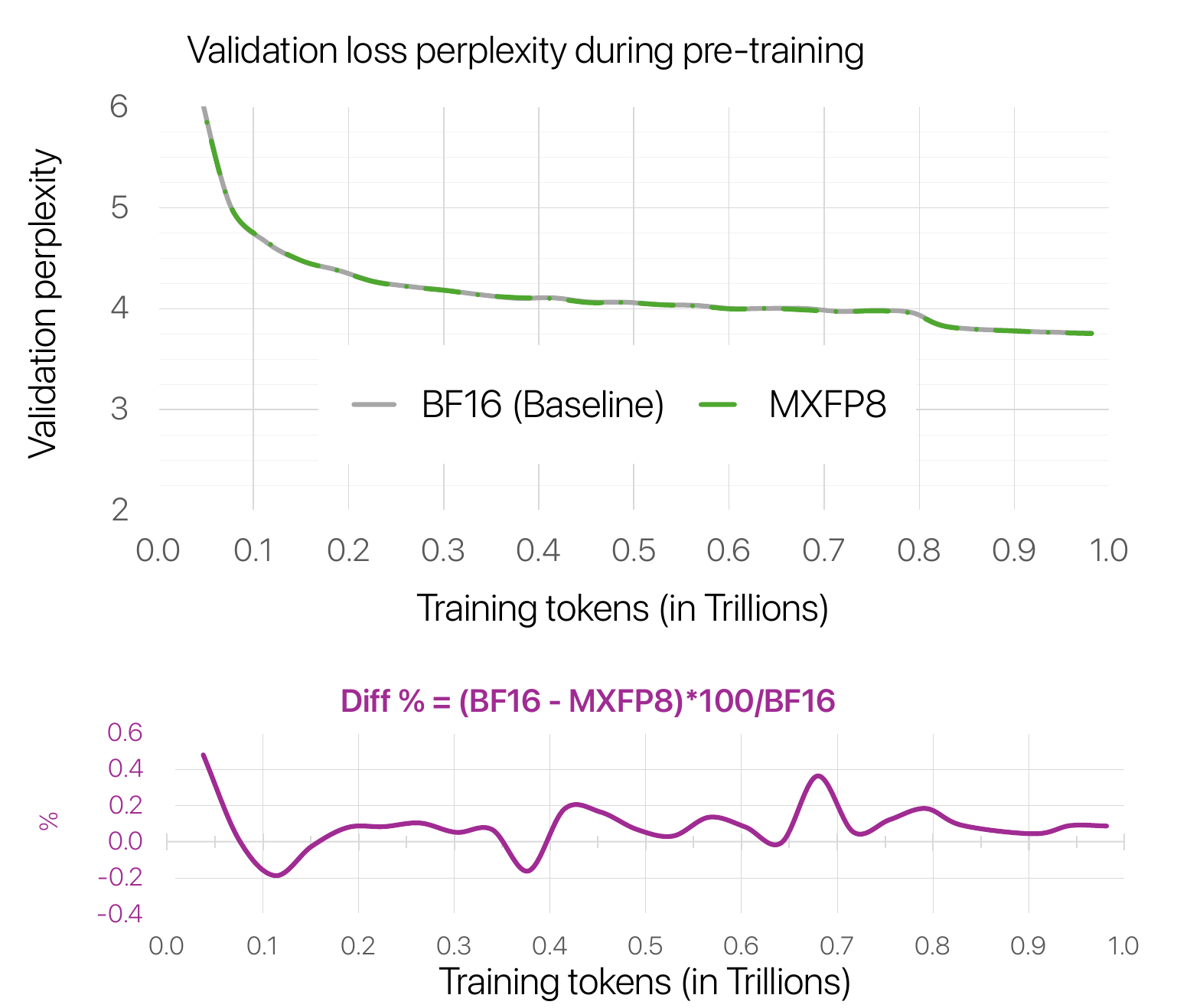}
    \caption{MXFP8 versus  \texttt{BF16}  for a MoE model} 
     \label{fig:moe}
\end{figure*}

Section~\ref{sec:results_mxfp8} discusses empirical data that shows \texttt{MXFP8} matches \texttt{BF16} accuracy (both training loss as well as downstream task accuracy). Transformer based mixture-of-experts (MoE) models are popular in literature. Figure~\ref{fig:moe} shows that \texttt{MXFP8} pre-training also matches \texttt{BF16} pre-training loss curve for a MoE setup that we experimented with. The MoE model has 16 billion total parameters and $\sim$2.5 billion active parameters and we train the model on 1 trillion tokens. We follow the same guidelines discussed in section~\ref{sec:dtype_tensors} for pre-training the MoE model. The pre-training phase uses a WSD~\cite{wen2024understandingwarmupstabledecaylearningrates} learning rate schedule. The final loss of the \texttt{MXFP8}-trained MoE model is within 0.1\% of \texttt{BF16}-training.

\subsection{Model configurations} 
\label{model_cfg}

We conduct numerical experiments on LLM pre-training with 
variants of Nemotron-4~\cite{parmar2024nemotron415btechnicalreport} models. 
Training and model details are described below. 
The 1T and 300B tokens dataset are a subset of the 17T data set discussed in~\cite{nvidia2024nemotron4340btechnicalreport}.
Table~\ref{tab:model_details} details
the parameters for the various models that were used.

\begin{table}[hb]
  \setlength{\tabcolsep}{5pt}
  \caption{Configuration for the LLM models.}
  \label{tab:model_details}
  \centering
  \begin{tabular}{llllllll}
    \toprule
    Model & Layers & Hidden & Attention & Sequence & Batch & Initial  & Final \\
    & & Size & Heads & Length & Size & LR & LR \\
    \midrule
    843M & 24  & 1024 & 16 & 4096 & 256 & 2.5e-4 & 2.5e-7 \\
    8B   & 32  & 4096 & 32 & 4096 & 1024 & 3e-4 & 3e-7 \\
    \bottomrule
  \end{tabular}
\end{table}

\end{document}